# Learning from Web: Review of Approaches

Vitaly Schetinin

## 1 Introduction

The symbolic machine learning techniques, the neural network approach, and genetic algorithms provide different methods of data analysis and knowledge discovery (Mooney, Shavlik, et al. 1989; Quinlan, J. 1994; Tank and Hopfield 1987; Craven and Shavlik 1993; Koza 1992). As these techniques have shown unique capabilities for analyzing symbolic and numeric data, we briefly survey underlying approaches that can be useful to solve a problem of learning from Web, which associated with a problem of knowledge discovery from large document collection, text categorization and document retrieval. Generally, the text categorization is the problem of identifying a set of predefined topics presented in a natural language document [Lewis 1992; Cohen 1995; Wiener et al. 1995; Wermter 1999]. Formally, this task is to output for each topic the probability that the topic is present. There are data-driven approaches to text categorization that applies neural networks to estimate the topic probability. Some categorization approaches have been concerned with constructing a "human readable" rule that render the operation of the system intuitively understandable, and heuristically adjustable. In contrast, the neural network is essentially nonlinear regression model for fitting high-order interactions in some feature space to binary topic assignments. These models are hard-to-understand, although they are hopefully high-performance.

There are several problems for learning the text document classification. One problem concerns with *over-fitting* which occurs when the learning system starts to memorize the training patterns, i.e., when it starts fitting the peculiarities of the training data, decreasing its performance on out-of-sample data (Wiener, Pedersen, and Weigend 1995; Schutze, Hill, and Pedersen 1995).

Other problems concern with document representation and the size of labeled example sample. Used for representing a document input features must be useful for document classification, i.e., performance of document classification is improved when each of features influences on a classification result (Lewis 1992). Statistical text learning algorithms can be trained to approximately classify documents on a sufficient set of labeled training examples. For these algorithms, key difficulty is that they require a large number of labeled training examples to learn accurately (Nigam, McCallum, et al. 1998). For example, a practical user of filtering system would prefer learning algorithms that can provide accurate classifications after hand labeling only a dozen documents rather than a thousand.

## 1.1 Symbolic Learning

The most promising symbolic machine learning technique for knowledge discovery is learning from example as a special case of inductive learning. This technique induces a general concept that best describes the positive and negative examples. Most of the symbolic learning algorithms produce the production rules or the concept hierarchies. These representation are ease to understand and efficient to perform. Among powerful algorithms for inductive learning are Quinlan's decision-tree building algorithms (Quinlan 1994) that take objects of a known class, described in terms of a fixed collection of attributes, and produce a decision tree that correctly classifies all the given objects. The algorithms minimize the expected number of tests to classify objects. Their results can be summarized as IF-THEN rules.

## 1.2 Neural Networks

In symbolic machine learning, knowledge is presented in the form of symbolic descriptions of learning concepts, e.g., production rules or concept hierarchies. In connectionist learning, knowledge is learned and remembered by a network of interconnected neurons, weighted synapses, and threshold logic units (Fisher and McKusick 1989; Craven and Shavlik 1996; Craven, and Shavlik 1997; Weigend, Huberman, and Rumelhart 1990). Learning algorithm are applied to adjust connection weights so that the network can correctly classify unknown examples. Among the neural networks, Backpropagation networks are most popular for their unique learning capability. Backpropagation networks are fully connected feed-forward models that consist of input, hidden, and output layers. Learning algorithm typically starts out with a random set of weights. The network adjusts its weights each time that it sees an instance, input-output pair. The weights connected to the output units are adjusted by a gradient descent method to reduce the errors. Backpopagation network updates its weights incrementally until the network stabilizes.

## 1.3 Genetic and Evolution Algorithms

Genetic algorithms were developed based on the principle of genetics using chromosomal operations such as crossover and mutation (Booker, Goldberg and Holland 1990; Goldberg 1989; Koza 1992). In these algorithms a population of individuals (potential solution) undergoes a sequence of unary (mutation) and higher order (crossover) transformation. After some number of generations, the algorithm converges, and the best individuals represent the desirable optimal solution.

Ivakhnenko (1995) suggested the genetic-based algorithms for self-organizing multi-layered neural networks that have minimal complexity. In other words, the networks trained by self-organization algorithm consist of minimal number of layers, neuron units and their interconnections. This approach allows the using unrepresentative training set and simultaneously it avoids a problem over-fitting of neural networks.

## 1.4 Hybrid Learning Techniques

There are studies that compared the performance of above-mentioned techniques for different applications. Mooney, Shavlik et al. (1989) found that symbolic learning algorithm ID3 was faster than Backpropagation network, but the last was more adaptive to noisy (contradictory and incorrect) data sets.

Fisher and McKusick (1989) found that using batch learning, Backpropagation performed as well as ID3 but it was more noise-resistant. Montana and Devis (1989) showed that using some domain-specific genetic operators to train the Backpropagation network, instead of using the conventional Backpropagation Delta learning rule, improved performance. Kitano (1990) are considered hybrid systems (genetic algorithms and neural networks) which performed symbolic induction using genetic algorithms and network, which is a fuzzy connectionist expert system. Other hybrid systems employ symbolic and neural net characteristics. Touretzky and Hinton (1988) and Gallant (1988) proposed connectionist production systems.

## 2 Document Representation

Words have properties, such as synonymy and polisemy that make them a less than ideal indexing language. These have motivated attempts to use more complex features extraction methods in text categorization and text retrieval. One strategy is available if a feature can be defined by the presence of two or more words in particular syntactic relationships (Feldman and Hirsh 1996). Another is to use cluster analysis or other statistical techniques to identify closely related features. Groups of such features can be replaced by a s single term corresponding to their logical and numeric sum.

Most systems have used words or words-couples for features. Lewis and Ringuette (1994) compared two categorization techniques, one a Bayesian classifier based on conditional probability model and other using the decision trees. The classifiers were separately constructed for each topic. The documents were presented as binary word vectors to substantially reduce the number of words in the representation. A different set of words was selected for predicting each topic using the information gain measure introduced to pick words with high individual predictive power. The information gain measure is based on the mutual information between the presence of a topic and the presence of a word over the documents in the corpus. It was showed the decision tree method gives higher performance than Bayesian classifier. Also the decision tree method resulted in a classifier whose rules were easier to interpret.

Apte et al. (1994) used a rule induction technique called SWAP-1 to produce disjunctive binary decision rules relating the presence of certain combinations of terms. As in the previous approach, separate classifiers were trained for each topic. The authors found that classification performance improved when word frequency value were used instead of binary presence/absence features, and that performance could also be improved by selecting for the local representation the most frequent words for a topic instead of the most informative words.

Wiener et al. (1995) used a document matrix containing word frequency information. The entries of each document vector, or a document profile, are computed as following $p_{dk} = \sqrt{f_{dk}}/\sqrt{\Sigma_{\text{vector}}(\sqrt{f_{di}})^2}$, where $f$ is word frequency. The square root dampens the effect of high counts and the normalization removes the effect of variations in document length. The authors also used a term selection aimed to find the subset of the original terms which seem the most useful for the classification task.

### 2.1 Term Selection

It is necessarily to select a set of terms that can adequately discriminate between given classes of documents while at the same time being small enough to serve as the features set for a classifier. Wiener et al. (1995) score all terms according to how well they serve as individual predictors of the topics. The score measures how unbalanced the term is across documents with and without topic

$r_k = \log((w_{tk}/d_t + 1/6)/(w^{\mathrm{T}}_{tk}/d_t + 1/6))$,

where $w_{tk}$ is the number of documents with the topic that contain the term, $d_t$ is the total number of documents with the topic. The authors found that about 20 terms yielded the best classification performance. Although the error on the training set can decrease by including more terms, classification performance on out-of-sample data quickly falls off after about 20 terms due to *over-fitting*.

Wiener et al. computed how well topics could be predicted using only a single key term. They surprisingly found that 20 of the 92 topics could be predicted with over 90% precision. Term selection has the advantage that relatively little computation is required and re-sulting features have direct interpretability. However, it is likely that many of the best predictors contain redundant information. To reduce the dimensionality of feature set, the authors applied latent semantic indexing (LSI) they improved by using global and local LSI representations. In particular, it was defined five broad meta-topics, and corpus was broke into five clusters, each of which containing all the documents. For these clusters, five cluster-directed LSI representations were computed.

## 2.2. FGEN Algorithm

Kudenko and Hirsh (1998) focused on the class of inductive learning methods used when a collection of text documents can be described as features vectors (i.e., as values assigned to a fixed set of attributes describing each object). The attributes can have the form of "*The subsequence S occurs somewhere in the sequence*", i.e., each attribute tests for presence or absence of a particular subsequence in the given sequence. A feature for every possible subsequence is created up to a certain length $k$ (using all $n$-grams for $n$ ranging between 1 and $k$). Such straightforward representation performed significantly worse than representation based on domain knowledge.

The method called FGEN creates a new feature as a kind of macro-feature that tests for the number of occurrences of each possible subsequence within a given sequence. FGEN representation was evaluated by comparing the accuracy of C4.5 (Quinlan 1994) and RIPPER (Cohen 1995). Results show that FGEN features often improved the accuracy of learning algorithms.

To represent the subsequences as feature vectors, the authors used positional and subsequence representations. In positional representation of subsequences each feature denotes a certain position of the sequence. The sequence ABB would be represented (P1 A) (P2 B) (P3 B), i.e., there is word A in position 1, word B in position 2, etc. However, such representation works only when all the data sequences are of the same length, since learners require that all data were described using identical sets of attributes. In alternative approach, it is considered the presence or number of occurrences of various subsequences in a sequence. These subsequences are generated for each $n$-gram that can occur in a sequence. Formally, FGEN feature can be written as a function $F$ from the set of all subsequences

$$F = s^{n1}_1 \wedge s^{n2}_2 \wedge \ldots \wedge s^{nk}_k,$$

where $s^{ni}_i$ called a minimum frequency restriction is a Boolean function whose value is *true* if subsequence $s_i$ occurs at least $n_i$ times in a sequence.

FGEN generates a feature set for each class separately and it returns the union of the set. The FGEN feature generation algorithm focuses on combination of subsequences that occur mostly in training examples of one class and not in the examples of the other class. Building features is incrementally done and it starts with a single example of the target class. After each iteration classification accuracy of new feature is evaluated on one third of train set. Generalization steps are continued while the new feature has a lower error rate than the feature in the previous step. In order to solve over-fitting problem, each new feature is generalized by removing a subsequence restrictions one by one from a feature. This is done by decrementing a minimum frequency restriction. In empirical tests on textual data, the length $n$-gram ranges between 2 and 5. FGEN representation improves accuracy across a wide range of domains.

## 2.3 Using Linguistic Phrases

Current learning algorithms of information extraction are able to provide a learner with features that capture some of the syntactic structure of natural language text. Such systems use syntactic heuristics to create linguistic patterns that can extract the desired information from the training documents. These patterns typically represent subject-verb or verb-direct relationships as well as prepositional phrase attachments. Fürnerkranz, Mitchell, and Riloff (1998) used a linguistic phrase as an input feature for text categorization problem that solved with a naive Bayes and RIPPER classifiers. The rule learner RIPPER is better able to focus on the small differences between similar pages in different classes. For Bayes classifier and RIPPER algorithms, only a few of the learned rules actually use the phrase features. Although phrasal features require more efforts to parse the syntax of test documents, they provide a better focus for rule learning algorithms. The obtained results show that phrasal features can improve the precision of learning algorithms.

## 2.4 Reducing Sample Size

Nigam, McCallum et al. (1998) describe algorithm that learns to classify text document more accurately by using unlabeled documents to augment the available labeled training examples. In experiments, the labeled document might be 10 postings the user judged as interesting or not. Learning algorithm can use the unlabeled postings available on UseNet. The technique reduces the need for labeled training examples by a factor three.

The unlabeled examples are able to boost learning accuracy because they provide information about the joint probabilities over words within the documents. For example, the document that contains word lecture tends to belong to the class of academic course web pages. Using this fact to classify unlabeled documents, we might find that the word homework frequently occurs in these documents that are believed to belong to the course class. Thus the co-occurrence of words lecture and homework over the large set of unlabeled examples provides useful information to construct a more accurate classifier that considers both lecture and homework as indicators of course class.

# 3 Symbolic Rule Learning

## 3.1 Bayes classifier

A number of authors (Lewis and Ringuette 1994; Craven, and Kumlien 1999) have used a naive Bayes classifier for text categorization. It estimates the probability that a document is a member of a certain class using probabilities of words occurring in documents of that class independent of their context.

The probability of document $d$ belonging to class $C$ is estimated by multiplying the prior probability $\Pr(C)$ of class $C$ with the product of the probabilities $\Pr(w_i|C)$ that the word $w_i$ occurs in documents of this class. This product is then normalized by the product of the prior probabilities $\Pr(w_i)$ of all words.

As many of the probabilities $\Pr(w_i|C)$ are typically 0.0, there is a related problem - many estimates of $\Pr(C|d)$ for the winning class tend to be close to 1.0 and often will be exactly 1.0 because of floating-point round-off-errors.

## 3.2 RIPPER algorithm

RIPPER (Cohen 1995; Cohen and Singer 1996; Cohen and Hirsh 1998) is rule learning algorithm based on the incremental reduced error-pruning algorithm. It learns single rules by greedily adding one condition at a time using information gain heuristic until the rule no longer makes incorrect predictions on the growing set, a randomly chosen subset of the training set. The learning rule is simplified by deleting conditions as long as the perform-ance of the rule does not decrease on the remaining set of examples. All examples covered by the resulting rule are then removed from the training set and a new rule is learned in the same way until all examples are covered by at least one rule.

The use of set-valued features makes RIPPER well suited for text categorization problems. A document is represented as a single set-valued feature that simply lists all the words oc-curring in the text. Note that for conventional learning algorithms, a document is typically represented as a set of binary features, encoding the presence or absence of a particular word in that document. This results in a very inefficient encoding of the training example because much space is wasted for specifying the absence of words in a document.

The phrasal and term clustering methods were investigated on Reuters collection. The phrasal indexing language consisted of simple noun phrases, i.e. head nouns and their im-mediate pre-modifiers. Functional words were removed from phrases. Phrases were formed using parts. The nearest neighbor clustering was used for clustering the similar features that define eight sets of meta-features. The meta-features were based on presence or absence of features in documents, or on the strength of association of features with categories of documents.

The surprise is the small number of features found to be optimal. For 14,704 and 1,300 training examples, peaks of 10 and 15 features respectively are smaller than one would ex-pect based on sample size. One possible trouble is over-fitting when a model trains on acci-dental as well as systematic relationships between feature value and category memberships. Effectiveness reaches a peak not much higher than that achieved on the unseen test set even if vary large feature set is used. Other possible explanation for the decrease in effectiveness with increasing feature set is that the probability of observing a word in document is inde-pendent of the probability of observing any other word in the document.

Cohen concluded that effectiveness of phrasal representational peaks at a much higher fea-ture set size (around 180 features) than that of a word-based representation. Maximum ef-fectiveness of the phrasal representation is substantially lower than that of the word-based representation. Term clustering did not significantly improve the quality of either a word-based or phrasal representation.

In work of Craven et al. (1998), the research efforts have been aimed to automatically creat-ing and maintaining a computing a computer-understandable knowledge base whose con-tent mirrors the WWW. The WWW knowledge base would consist of computer under-standable assertions in symbolic, probabilistic form. The approach explored is to develop a trainable system that can taught to extract various types of information by automatically browsing the Web.

Two data sets for the experiments were assembled, thirst is a set of 4,127 pages and hyper-links drawn four Computer Science Departments, second is a set of 4,120 pages from nu-merous other CS departments.

One task for system is to identify new instances of hierarchical classes from the text sources from the Web. Statistical page-classification method is a probabilistic model of each class using labeled training data. Probabilistic models, called bag-of-words ones, ig-nore the sequence in which the words occur. For classifying Web pages, the naive Bayes method was modified and used. The authors have found that the classification to be more accurate when using a restricted up to 2000 words vocabulary.

The models learn to take into account features as the pattern of connectivity around a given page, or words occurring in neighboring pages. For example, it learns a rule of the form "A page is a *Course* home page if it contains the word *textbook* and is linked to a page that contains the word *assignment*". Such rules are called the first-order ones. The learning algorithm that is able to induce them is greedy covering algorithm for learning the function-free Horn clauses. This algorithm can represent a background relation as, for example, *link_to(Page, Page)*. The first argument is the page on which the link occurs, and second is the page to which it is linked.

Along with each predicting instances, it is calculated an associated measure of confidence. The confidence is determined as an estimate of the error of the clause making the predictions.

The authors hypothesize that relations among class instances are often represented as *hyperlink paths* in the Web. To recognize relation instances, they used above algorithm for learning and found that a hill-climbing strategy is unable to learn rules for paths consisting of more than one hyperlinks. The search process was used that consists of two phases. In the first phase, the path part of the clause is learned, and in second, additional literals are added to the clause. The presented approaches take advantages of the special structure of hypertext by considering relationships among Web pages, their hyperlinks, and specific words.

## 4 Neural Networks

Neural networks provide a convenient knowledge representation for document retrieval in which nodes typically represent objects such as keywords, authors, and index terms and links represent their weighted associations (of relevance). Learning property of Back-propagation networks and the parallel search property of the Hopfield network provide effective means for identifying relevant information.

The model of Belew (1989) is probably one of first connectionist models adopted for document retrieval. His model is tree-layer neural network of authors, index terms, and documents. The system used relevance feedback from its users to change its representation of these attributes.

Instead of using a three-layer network, Chen's systems (1992; 1993) generated an automatically created weighted network of keywords from large textual databases and integrated it with several thesauri man made. These systems developed a single-layer network of keywords (concepts).

## 4.1 Hopfield Networks

Knowledge can be stored in single-layered interconnected neurons (nodes) and weighted synapses (links) and it can be retrieved based on the networks' parallel relaxation method (Tank and Hopfield 1987). In Chen's systems, implementation incorporated the basic Hopfield net iteration and convergence ideas. A step of algorithm follows.

*First, assign the synaptic weights*: The resulting links represent probabilistic, synaptic weights between any two concepts from thesauri that were generated automatically using a similarity function. The synaptic weight from node $i$ to node $j$ represents as $t_{ij}$.

*Second, initialize the search term*: An initial set of search terms serves as an input pattern. Each node in the network that matches the search terms is initialized at time 0 by a weight of 1.

$\mu_i(0) = x_i$, $i = 1, \ldots, n$,

where $\mu_i(t)$ is the output of node $i$ at time $t$; $x_i$ is input pattern ranged from 0 to 1 for node $i$;

*Third, iterative compute the weights*: Each newly activated node derives its new weights by summarizing the products of weights assigned to its neighbors and their synapses.

$\mu_i(t + 1) = f[\Sigma t_{ij} \mu_i(t)]$, $i, j = 1, \ldots, n$,

$f(net_j) = 1/[1 + \exp(-p)]$, $p = (net_j - \vartheta_j)/\vartheta_0$,

where $f$ is sigmoid function; $net_j = \Sigma t_{ij} \mu_i(t)$; $\vartheta_j$ serves as a threshold or bias, and $\vartheta_0$ is used to modify the shape of sigmoid function.

*Fourth, check on convergence*: Above process is repeated, while a change in terms of outputs between two iterations is more than a given small value $\varepsilon > 0$.

## 4.2 Multilayered Neural Networks

Wiener et al. (1995) used a neural network approach to text categorization problem. For linear analysis, the logistic regression, which is appropriate for modeling binary output variables, was used. Its functional form is

$p = 1/(1 + e^{-\eta})$,

where $\eta = \beta x$ is a linear combination of input features. The logistic function guarantees that $p \in (0, 1)$. Logistic regression can be converted into a binary classification method by comparing the output probabilities with the thresholds.

Nonlinear neural network classifiers were used to extend logistic regression by modeling higher-order term interaction between features.

Major components of a neural network are architecture (e.g. unit connectivity, and activation functions) and search algorithm. The search is performed in the training process to find a set of weights, which is able to minimize the cost function. As a search technique, the back-propagation method of gradient descent in weight space was used. Applying logistic regression allows controlling the fitting process by using both additional priors to the cost function and early stopping in training.

The simplest linear classifier network has an output unit with a logistic activation and no hidden layer. To model nonlinear relationship between the input and output variables, the one or more hidden layers of nonlinear activation functions are used.

Two different network architectures were explored, one flat and other modular. In the first case, the entire training set and global LSI representation were used to train a separate network for each topic.

## 4.3 Over-fitting Problem

Major problem for all topics is yet over-fitting a network. While a network can be learned with extremely low error, the error on a separate validation set typically reaches its minimum and then begins rising. To alleviate this problem, a simple regularization scheme based on weight-elimination was applied. In this scheme authors add a term penalizing network complexity to the cross-entropy cost function. The modular approach allowed de-composing the learning problem into a set of smaller problems. In first step, a network trained on full training set to estimate the probability that each of the five meta-topics is present in the document. In second step, a set of five network groups corresponding to the meta-topics. Each group consists of a separate network for each topic in the group, trained only on the region of the corpus corresponding to the meta-topic. The meta-topic network uses fifteen hidden units, and local topic networks use six hidden units. Meta-topic network determines on a high-level the incoming documents that can be processed by the local networks.

A relatively high error can arise if the example are perfectly separated but the probability estimates are mid-range rather than near 0 and 1. The binary decisions are forced by applying the thresholds. The decision thresholds are picked in one of two ways suggested.

Wiener et al. concluded that, although, the nonlinear networks are better than the linear models, the difference in their performance is very slight. Possibly, there are too few positive examples to support nonlinear fits, which generalize well. The nonlinear networks are unable to extract features, which can generalize the out-of-sample data before training is halted with early stopping.

## 4.4 Comparative Analysis

A routing system (Schutze, Hill, and Pedersen 1995) uses a query and a large sample of documents that have been identified as relevant or not relevant to construct a classification rule that ranks unlabeled documents according to their likelihood of relevance. The documents have to be assigned to one of two categories, relevant or non-relevant. A central problem in routing is the high dimensionality of the features space, typically hundreds of thousands. One solution is to reduce dimensionality by using subsets of the original features or transforming them. Another one is does not reduce dimensionality, but employs a learning algorithm without explicit error minimization. An example of such an approach is relevance feedback via query expansion (QE). Two different forms of dimensionality, LSI and optimal term selection are investigated to evaluate which form is most effective for routing problem. The authors examined a number of different methods: QE, linear discriminant analysis (LDA), logistic regression (LR), both linear and nonlinear neural network. The Tipster collection and the TREC routing tasks were used to test classifiers and representation.

LSI is technique that represents features and documents by a low-dimensional linear combination of orthogonal indexing variables (). LSI captures the theme a document by analyzing the patterns of occurrence between terms. In contrast, a term-based representation scheme hides thematic similarity of documents. This makes it difficult to obtain an accurate measurement of relevance. LSI avoids this problem by representing the theme of a document rather than specific terms.

The authors used LSI that differs in two aspects. First, it was applied local LSI to a region of the document space that is in the neighborhood of the query. Second, LSI representations are used as input parameters to a learning algorithm. LSI applies matrix decomposition to the term by document matrix of the collection to generate a large number of orthogonal LSI factors. A small number of the most important factors are then selected to approximate the covariance structure of the full collection. While a sparse algorithm does not need to calculate all orthogonal factors, it is still difficult to compute the LSI representation for the TREC collection, since it contains over a million terms and documents. The authors used only 100 most important factors in experiments. The majority of previous algorithms require that the number of features be restricted in some way (i.e. a set of features is not optimized in learning process).

A neural network consisting of input and output units was trained by back-propagation method. The neural networks have the ability to fit a wide range of distribution documents, for example, any member of the exponential family can be modeled. However, this capacity often leads to the danger of over-fitting, when the trained network fits the training data too precisely and does not generalize the full population. To protect network against over-fitting, the authors used a validation set which contains one thirds of full training data. When the model parameters are updated in iterative manner, the error on validation set is computed. Training continues until the validation error goes up, which indicates that over-fitting has set on. This procedure defines the number $n$ of iterations that improve generalization. In final, the model parameters are recomputed by training on the entire training set for n iterations. Computational expense of this procedure is less than systematic cross-validation.

The linear network consists only of input and output units. For units of the network, the sigmoid was chose as the activation function. Linear neural network and logistic regression have the same probabilistic model, but combination with gradient descent is better suited to void over-fitting. A non-linear network has hidden units, which can be interpreted as features detectors that estimate the probability of a feature being present in the input.

The reduced representational scheme substantially decreases training time and is less prone to over-fitting because there are fewer parameters. For experiments, the Tipster corpus was used that consists of 3.3 gigabytes of text in over one million documents. This corpus was preprocessed (e.g. performing document parsing, stemming a word, and removal of terms from the stop word list). The terms consisted of single words and two-word phrases (defined as an adjacent word pair) that occur over five times in the corpus. It was produced over 2.5 million terms.

The most important conclusion is first that the linear neural networks work better than logistic regression. This indicates that the logistic model is over-fitting the training data because they are using exactly the same model. The second, the experimental results suggest that there is not advantage to adding nonlinear components (i.e. hidden units) to the neural network, probably because there is not enough information in the Tipster data collection to accurately learn complex models. The third, if LSI is more effective for classification techniques such as LDA and LR, which have no protection against over-fitting, the neural networks perform equally well with either set of features. A classification technique suffers from over-fitting when it improves performance over the training documents but reduces performance when applied to new documents.

## 4.3 Kohonen Map

Other researcher used different neural networks for text mining. A Kohonen network, which produced a two-dimensional grid representation for $N$-dimensional features, was adopted for document retrieval (Honkela et al. 1995; Kaski et al. 1998; Kohonen 1998). Kohonen's feature map was applied to construct a self-organizing, unsupervised learning, visual representation of the semantic relationships between input documents.

## 5 Genetic Approaches to Document Retrieval

Gordon (1988) developed a genetic algorithm for document retrieval. The descriptions (keywords) of competing document are associated with a document and altered by using genetic mutation and crossover operators. A keyword represents a gene (a bit pattern), a document's list of keywords represents individuals (a bit string), and a collection of documents a user initially judged represents the initial population. By using a matching function (fitness measure), the initial population evolved trough a generation and eventually con-verged to an optimal population - a set of keywords that best described the documents. Petry et al. (1993) applied genetic programming to a Boolean query system modified in order to improve classification precision. They explore the effect of adopting genetic algorithms, the impact of genetic operators, and algorithm searching capabilities.

## 5.1 Genetic Algorithm

A genetic algorithm maintain a population of individuals, $P(t) = x_1, \ldots, x_n$ at iteration $t$. Each individual represents a potential solution to the problem it is implemented as some complex data structure $S$. The solutions $x_i$ are evaluated by some measure of fitness. Then a new population at iteration $t + 1$ id formed by selecting the fitter individuals. Some individuals of the new population are transformed by using the genetic operations to form new solution. Transformations $m_i$ of mutation type create new individuals by a small change in an individuals by combining parts from several individuals. The control parameters for genetic operations must be carefully selected to obtain better performance. After some number of iteration, the program converges - the best individuals represent the optimal solution. The step of genetic algorithm typically follows.

*First, initialize the population*: We need to predetermine both the number $m$ of genes for each individual and the total number of chromosomes, *popsize*, in the initial population. Each gene (bit) in the chromosome (bit string) represents a certain keyword or concept. A gene decides the existence (1) or absence (0) of a concept. The initial population contains a set of documents, which were judged relevant by a user. The goal of genetic algorithm was to find an optimal set of documents which best matched the searcher's request.

*Second, reproduce an individual*: Reproduction is the selection of a new population based on the fitness values. Fitter individuals have better chances to be selected for reproduction. We can assume that slots of a roulette wheel sized by the total fitness of population. Then each chromosome has a certain number of slots proportional to its fitness value. Spinning the wheel *popsize* times, each time we select a single chromosome for a new population.

*Third, recombine a chromosome*: Initially, we predetermine a probability $p_c$ in order to define the expected number $p_c \times popsize$ of chromosomes which should be undergo the crossover operation. Pairs of chromosomes, which were randomly selected, are coupled. For each pairs we generate a random number *pos* ranged from 1 to $m$. Further, coupled chromosomes exchange genes at the crossing point *pos*. The next operator, mutation, was performed on bit-by-bit basis. The mutation probability $p_m$ defines the expected number of mutated bits $p_m \times m \times popsize$. Every bit in all chromosomes has an equal chance to undergo mutation, i.e., change from 0 to 1 or vice versa. Typically, $p_c$ predetermined ranging between 0.7 and 0.9, and $p_m$ ranging between 0.01 and 0.03.

*Fourth, check on convergence*: Following reproduction, crossover and mutation, the next population was generated. Above steps were repeated until the system reached a predetermined number of generations or converged (no improvement in the overall fitness of the population).

## 6 Hybrid Learning from Web

Rose and Belew (1991) extended three-layered neural network to a hybrid connectionist and symbolic system called SCALIR, which used analogical reasoning to find relevant documents.

Wermter et al. (1999) developed and examined neural/symbolic agents for text routing on the Internet. Some information retrieval strategies were integrated with non-linear recurrent neural networks such as simple recurrent networks and recurrent plausibility networks that are able to consider the word order (McGarry, Wermter and MacIntyre ). The recurrent plausibility networks are extended with a short-term memory to be useful to process the text sequences in a robust manner. The network with $n$ hidden layers processes both the current input and the incremental context from the $n$-1 previous time steps. The input to a hidden layer $L_n$ is constrained by the underlying layer $L_{n-1}$ as well as the incremental context layer $C_n$. For computing the activation of unit $L_{ni}(t)$ at time $t$, it is used the weighted activation of previous layer $L_{(n-1)i}(t)$ units and the units in current context of this layer $C_{ni}(t)$ limited by the logistic function $f$

$L_{ni}(t) = f(\sum w_{ki}L_{(n-1)i}(t) + \sum w_{li}C_{ni}(t)).$

The units in the context layers perform a time-averaging of the information using the hysteresis value $\varphi_n$

$C_{ni}(t) = (1 - \varphi_n)L_{ni}(t - 1) + \varphi_n C_{ni}(t - 1).$

The hysteresis value $\varphi_{(n-1)}$ is lower than value $\varphi_n$ in the next context layer $C_n$. This ensures that the context layer, which is closer to the input layer, will perform as memory that represents a more dynamical context.

The recall and precision rates for recurrent networks trained on the Reuters newswire corpus were fairly high given the degree of ambiguity. The performance of unseen news titles has been even better that the performance on the training data. This convincingly demonstrates that over-fitting of the neural networks on the training set does not exist.

# 7 Conclusion

In reality, user-provided relevance feedback information may be limited in quantity and noisy (i.e., contradictory or incorrect). Neural network approach, as the empirical evidences have demonstrated, has document noise-resistant capability. For large-scale applications, the recurrent neural networks and some genetic algorithms may suffer from extensive computation time, over-fitting and lack of interpretable results.

On the other hand, symbolic learning effectively produced simple production rules or decision tree representations. The user's acceptance of the analytical results that an intelligent system provides remains to be determined.

First, we can conclude that a training set, which to be large collection of labeled documents, causes key difficulty of statistical approach to learning rules. This difficulty appears because a user must manually label the thousand documents to deploy them on different categories. We can expect that application of the self-organizing neural network (Ivakhnenko et al. 1995) modified to learn a text categorization rule could significantly reduce length of labeled training set. This hypothesis is based on that algorithm for self-organization decomposes a complex rule on a number of simplest rules, which can be trained on more short training set. Although training set will be shorter, classification accuracy will not be decreased. Moreover, classification accuracy can even increase.

Second, we know, that over-fitting problem appears when a network that was trained with more number of input features, interconnections, and neuron units has lower accuracy on testing set than a trained network that uses a less number of them. There is a pick of accuracy that points to network of optimal complexity (optimal number of features, network's interconnections, and units). We can expect that algorithm of self-organization will able to find the pick that points to optimal complexity of neural network trained for text categorization.

## References

1. Apte C. et al., 1994. Automated learning of decision rules for text categorization. ACM Transaction on Information Systems, vol. 12, 233-251.

2. Belew R. 1989. Adaptive information retrieval. In Proceedings of the 12th Annual Informational ACM/SIGIR Conference on Research and Development in Information Retrieval, 11-20, Cambridge.

3. Booker L., Goldberg D., and Holland J. 1990. Classifier systems and genetic algorithms. In Machine Learning, Paradigms and Methods, 235-282, MIT Press.

4. Chen H., and Lynch K. 1992. Automatic construction of networks of concepts characterizing document databases. IEE Transactions on Systems, Man and Cybernetics, vol. 22, 885-902.

5. Chen H., Lynch K. at al. 1993. Generating, integrating and activation thesauri for concept-based document retrieval. IEE EXPERTS, Special Series on Artificial Intelligence in Text-Based Information Systems, vol. 8, 25-34.

6. Cohen W. 1995. Text categorization and relational learning. In Proceedings of the International Machine Learning Conference (ICML-95), 124-132, Morgan Kaufmann.

7. Cohen W., and Singer Y. 1996. Context-sensitive learning methods for text categorization. In Proceedings of the 19th Annual International ACM Conference on Research and Development in Information Retrieval, 307-315.

8. Cohen W., and Hirsh H. 1998. Joins that generalize: text classification using WHIRL. In Proceedings of the Fourth International Conference on Knowledge Discovery and Data Mining.

9. Craven M., Shavlik J. 1993. Learning symbolic rules using artificial networks, In Proceedings of the 10th Int. Conference on the Machine Learning:.

10. Craven M., and Shavlik J. 1996. Extracting tree-structured representation of trained networks. In Touretzky D., Mozer M., and Hasselmo M., eds., Advances in Neural Information Processing Systems, vol. 8, 24-30.

11. Craven M., Shavlik J. 1997. Using neural networks for data mining, Future Generation Computer Systems: Special Issue on Data Mining.

12. Craven M., DiPasquo D., et al. 1998. Learning to extract symbolic knowledge from the WWW. In Proceedings of 15th National Conference on Artificial Intelligence (AAAI-98).

13. Craven M., and Kumlien J. 1999. Constructing biological knowledge bases by extracting information from text sources. In Proceedings of the International Conference on Intelligent Systems for Molecular Biology (ISMB-99).

14. Feldman R., and Hirsh H. 1996. Mining associations in text in the presence of background knowledge. In Proceedings of the 2th International Conference on Knowledge Discovery from Databases.

15. Fisher D., and McKusick K. 1989. An empirical comparison of ID3 and backpropagation. In Proceedings of the 12th International Joint Conference on Artificial Intelligence (IJCAI-89), 788-793.

16. Fürnerkranz J., Mitchell T., and Riloff E. 1998. A case study in using linguistic phrases for text categorization on the WWW. Working Notes of the 1998 AAAI/ICML Workshop on Learning for Text Categorization

17. Goldberg D. 1989. Genetic Algorithms in Search, Optimization, and Machine Learning. Addison-Wesley Reading, MA.

18. Gallant S. 1988. Connectionist expert system. Communications of the ACM, vol. 31, 152-169.

19. Gordon M. 1988. Probabilistic and genetic algorithms for document retrieval. Communications of the ACNM, vol. 31, 1208-1218.

20. Honkela T., Pulkki V., and Kohonen T.1995. Contextual relation of words in Grimm tales, anylized by self-organizing map. In Proceedings of International Conference on Artificial Neural Networks, ICANN-95, Paris, 3-7.

21. Kaski S., Lagus K., Honkela K., and Kohonen T.. 1998. Statistical aspects of the WEBSOM system in organizing document collections. Computing Science and Statistics, (Scott, D. W., ed.), Interface Foundation of North America, Inc.: Fairfax Station, VA, 29, 281-290.

22. Kitano H. 1990. Empirical studies on the speed of convergence of neural network training using genetic algorithms. In Proceedings of the 8[th] National Conference on Artificial (AAAI-90), 789-795, Boston.

23. Kohonen T. 1998. Self-organization of very large document collections: state of the art. In Proceedings of ICANN98, the 8[th] International Conference on Artificial Neural Networks, vol. 1, Springer, London, 65-74.

24. Koza J. 1992. Genetic Programming: On the Programming of Computers by Means of Natural Selection. The MIT Press, Cambridge, MA.

25. Kudenko D. and Hirsh H. 1998. Feature generation for sequence categorization. In proceedings of the AAAI-98.

26. Ivakhnenko A.G., Ivakhnenko G.A., and Müller J.A. 1995. Self-Organization of Neural Networks with Active Neurons, Pattern Recognition and Image Analysis, vol. 4, 185-196.

27. Lewis D., and Ringuette M. 1994. A comparison of two learning algorithms for text categorization. In Symposium on Document Analysis and Information Retrieval, 81-93.

28. Lewis D. 1992. Feature selection and feature extraction for text categorization. In Proceedings of Speech and Natural Language Workshop, 212-217, Morgan Kaufmann.

29. McGarry K., Wermter S., and MacIntyre J. Hybrid neural syatems: from simple coupling to fully integrated neural networks. Neural Computing Surveys, http://www.icsi.berceley.edu/~jagota/NCS.

30. Montana D., and Davis L. 1989. Training feed-forward neural networks using genetic algorithms. In Proceedings of the Eleventh International Joint Conference on Artificial Intelligence (IJCAI-89), 762-767, MI.

31. Mooney R., Shavlik J., et al. 1989. An experimental comparison of symbolic and connectionist learning algorithms. In Proceedings of the 12[th] International Joint Conference on Artificial Intelligence (IJCAI-89), 775-780.

32. Nigam K., McCallum A., et al. 1998. Learning to classify text from labeled and unlabeled documents. In Proceedings of the 15[th] National Conference on Artificial Intelligence (AAAI-98).

33. Petry F., Buckles B. et al. 1993. Fuzzy information retrieval using genetic algorithms ad relevance feedback. In Proceedings of the ASIS Annual Meeting, 122-124.

34. Quinlan, J. 1994. C4.5: Program for Machine Learning. San Mateo, CA: Morgan Kaufmann.

35. Rose D., and Belew R. 1991. A connectionist and symbolic hybrid for improving legal research. Int. Journal of Man-Machine Studies, vol. 35, 1-33.

36. Schutze H., Hill D., and Pedersen J. 1995. A comparison of classifiers and document representation for the routing problem. Proceedings of SIGIR'95, 229-237, ACM Press.

37. Tank D., and Hopfield J. 1987. Collective computation in neuron-like circuits. Scientific American, vol. 257, 104-114.

38. Touretzky D., and Hinton G. 1988. A distributed connectionist production system. Cognitive Science, vol. 12, 423-466.

39. Weigend A. S., Huberman B. A., and Rumelhart D. E. 1990. Predicting the future: a connectionist approach. International Journal of Neural Systems, vol. 3, 193-209.

40. Wermter S. 1999. Hybrid neural plausibility networks for news agents. In print.

41. Wiener E., Pedersen J., and Weigend A. 1995. A neural network approach to topic spotting. In Proceedings of the 4[th] annual symposium on Document Analysis and Information Retrieval (SDAR-95), 317-321.